\documentclass[runningheads]{llncs}
\usepackage{graphicx}

\usepackage{tikz}
\usepackage{comment}
\usepackage{amsmath,amssymb}
\usepackage{color}
\usepackage{adjustbox}

\begin{document}

\pagestyle{headings}
\mainmatter
\def\ECCVSubNumber{1381}

\title{Semi-supervised Learning with a Teacher-student Network for Generalized Attribute Prediction}

\titlerunning{Semi-supervised Learning for Generalized Attribute Prediction}

\author{Minchul Shin \inst{1}\orcidID{0000-0002-0638-2017}}

\authorrunning{M. Shin.}

\institute{Search Solutions Inc., Republic of Korea \\
\email{min.stellastra@gmail.com}}

\maketitle

\begin{abstract}
This paper presents a study on semi-supervised learning to solve the visual attribute prediction problem. In many applications of vision algorithms, the precise recognition of visual attributes of objects is important but still challenging. This is because defining a class hierarchy of attributes is ambiguous, so training data inevitably suffer from class imbalance and label sparsity, leading to a lack of effective annotations. An intuitive solution is to find a method to effectively learn image representations by utilizing unlabeled images. With that in mind, we propose a multi-teacher-single-student (MTSS) approach inspired by the multi-task learning and the distillation of semi-supervised learning. Our MTSS learns task-specific domain experts called teacher networks using the label embedding technique and learns a unified model called a student network by forcing a model to mimic the distributions learned by domain experts. Our experiments demonstrate that our method not only achieves competitive performance on various benchmarks for fashion attribute prediction, but also improves robustness and cross-domain adaptability for unseen domains.

\end{abstract}

\section{Introduction}
Visual attributes are qualities of objects that facilitate the cognitive processes of human beings. Therefore, predicting the attributes of an object accurately has many useful applications in the real world. For example, a search engine can use predictions to screen products with undesirable attributes instead of using noisy metadata provided by anonymous sellers~\cite{adhikari2019progressive}. Attribute prediction is essentially a multi-label classification problem that aims to determine if an image contains certain attributes (\textit{e.g., colors and patterns}). However, attribute prediction is known as a very challenging task based on the expense of annotation, difficulty in defining a class hierarchy, and simultaneous appearance of multiple attributes in objects. Although recent works have shown competitive results on various benchmark datasets~\cite{chen2019multi,guo2019visual}, we identified several additional conditions that must be satisfied to solve the aforementioned issues for such methods to be useful in real-world applications: domain-agnostic training, the use of unlabeled data, and robustness/generalization.

\begin{figure}[t]
	\begin{center}
		\includegraphics[width=0.75\linewidth]{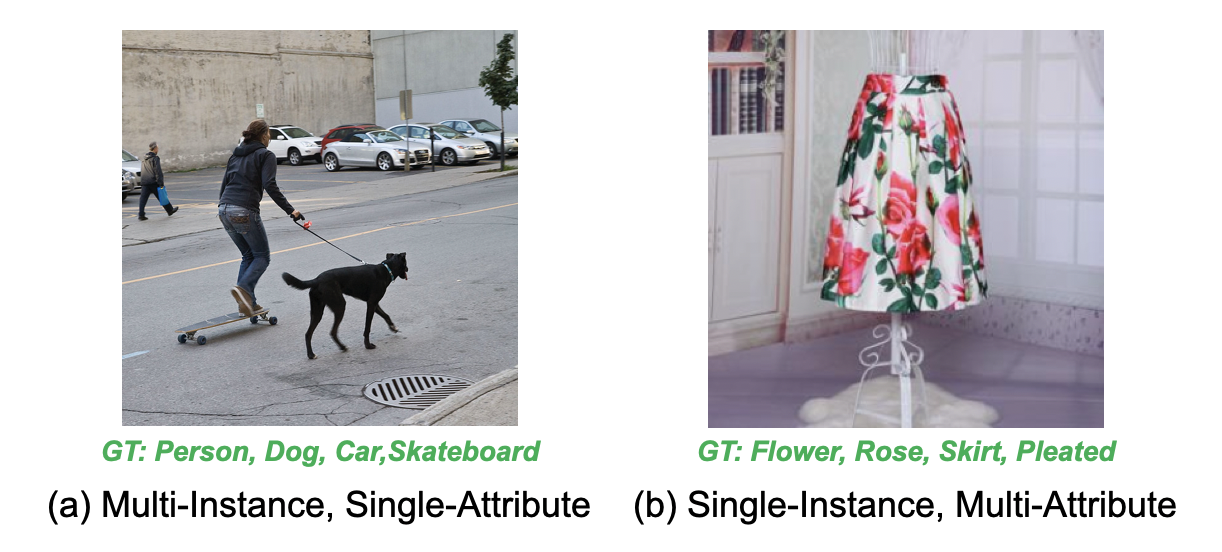}
	\end{center}
	\caption{Two different types of visual attribute prediction tasks. Image (a) was sampled from MSCOCO~\cite{lin2014microsoft} and image (b) was sampled from DeepFashion~\cite{liu2016deepfashion}.}
	\label{fig:two_types}
\end{figure}

\noindent
\textbf{Use of unlabeled data.} The human perception of an attribute is intrinsically subjective, leading to sparse and ambiguous annotations in datasets. Additionally, attributes have very-long-tailed distributions, leading to severe class imbalance (\textit{e.g., pattern: solid versus zebra}). Therefore, even if the total number of annotated images is large, models can suffer from extreme shortages of usable training data following class balancing in many cases~\cite{dal2015undersampling}. An intuitive solution is to use unlabeled images for training in a semi-supervised manner to enhance the ability of a model to represent the visual information included in images. The teacher-student paradigm proposed in this paper facilitates this process with no preprocessing tricks, such as obtaining pseudo-labels based on predictions~\cite{cevikalp2019semi,yalniz2019billion,xie2019self}.

\noindent
\textbf{Robustness/generalization.} Robustness measures how stable and reliable a network is when making decisions in the presence of unexpected perturbations in inputs~\cite{zheng2016improving}. Generalization measures how well a trained model performs with a domain shift or if a model can be generalized sufficiently with insufficient training data~\cite{kawaguchi2017generalization,neyshabur2017exploring}. In addition to high accuracy on benchmarks, these aspects must also be considered as crucial factors for measuring model quality.

\noindent
\textbf{Domain-agnostic training.} To achieve state-of-the-art performance on benchmarks, recent works~\cite{zhang2019task,liu2016deepfashion,liu2018deep,arslan2019multimodal} have relied on domain-specific auxiliary information during training. However, such methods have obvious limitations in terms of expansion to additional domains, which is an important feature for real applications. Therefore, it is preferable to avoid using any domain-specific information during training. We refer to this principle as domain-agnostic training.

\noindent
\textbf{Main contributions.} In this paper, we propose a multi-teacher-single-student (MTSS) method for generalized visual attribute prediction that aims to train a single unified model to predict all different attributes in a single forward operation. This paper makes the following main contributions. First, we introduce an MTSS method that learns from multiple domain experts in a semi-supervised manner utilizing unlabeled images. We also show the advantages of label embedding for training teacher networks, which are used as domain experts. Second, we demonstrate that the MTSS method can enhance model quality in terms of robustness and generalization without sacrificing benchmark performance by learning from images in distinct domains. Finally, our approach is fully domain agnostic, meaning it requires no domain-specific auxiliary supervision, such as landmarks, pose detection, or text description, for training. By learning using only attribute labels, our model not only outperforms previous methods under the same conditions~\cite{corbiere2017leveraging,chen2012describing,huang2015cross}, but also achieves competitive results relative to existing domain-specific methods~\cite{liu2016deepfashion,wang2018attentive,quintino2019pose,liu2018deep} that use additional auxiliary labels for supervision.

\begin{figure}[t]
	\begin{center}
		\includegraphics[width=0.80\linewidth]{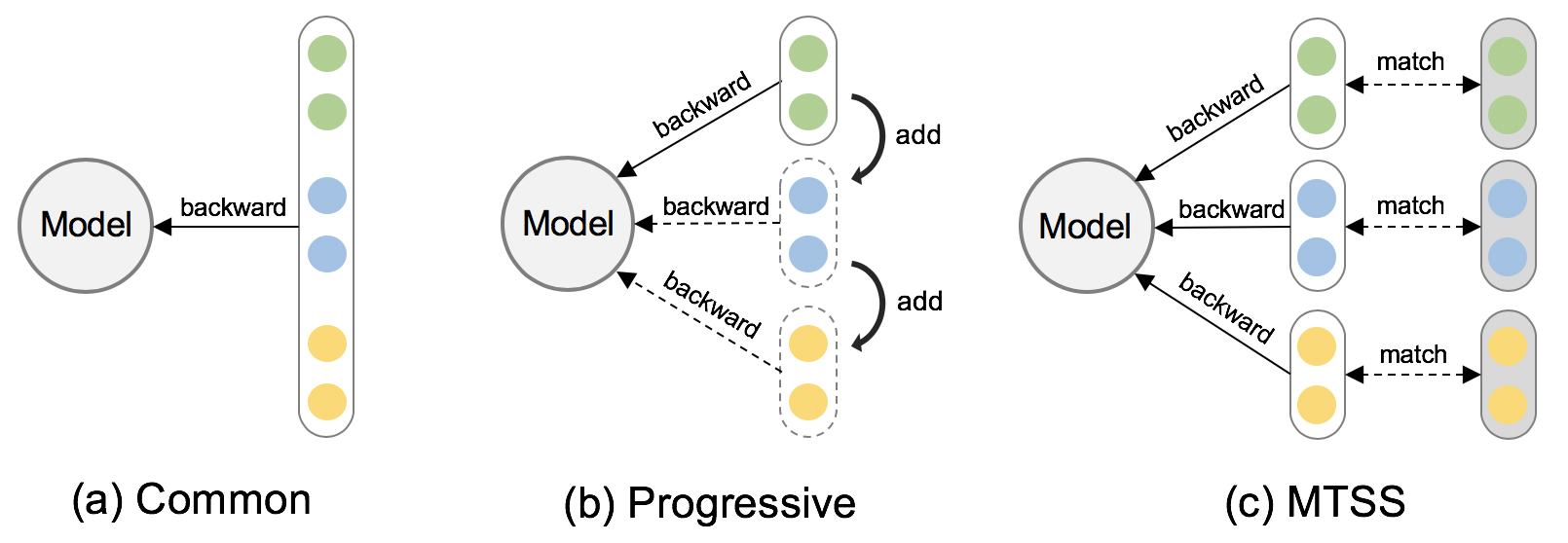}
	\end{center}
	\caption{Conceptual summary of previous methods and the proposed method denoted, as common~\cite{zhang2019task,liu2016deepfashion,liu2018deep,arslan2019multimodal}, progressive~\cite{adhikari2019progressive}, and MTSS.}
	\label{fig:concept}
\end{figure}

\section{Related Work}
\noindent
\textbf{Visual Attributes.}
Visual attribute prediction (VAP) is a multi-label classification task that has been widely studied~\cite{zhang2019task,arslan2019multimodal,adhikari2019progressive,liu2018deep,wang2018attentive,liu2016deepfashion,quintino2019pose,park2019study}. There are two main types of VAP tasks, as illustrated in Figure~\ref{fig:two_types}. We are particularly interested in type (b), which is frequently observed in the fashion domain. This is because our interests lie in the ability of a model to predict multiple simultaneously appearing attributes, rather than localizing interest regions, which is a matter of detection. The attributes appearing in type (b) are not necessarily related to visual similarity. For example, they may only be related to low-level characteristics, such as color, which makes the task more interesting. In the fashion domain, some works~\cite{zhang2019task,liu2016deepfashion,liu2018deep,arslan2019multimodal} have used landmarks for clothing items, pose detection, or textual item descriptions to improve overall accuracy. However, such strong requirements regarding auxiliary information limit such methods to domain-specific solutions. In contrast, some works have focused on attention mechanisms~\cite{wang2018attentive,quintino2019pose,zhang2019task}. Zhang \textit{et al.}~\cite{zhang2019task} proposed a task-aware attention mechanism that considers the locality of clothing attributes according to different tasks. Wang \textit{et al.}~\cite{wang2018attentive} proposed landmark-aware attention and category-driven attention. The former focuses on the functional parts of clothing items and the latter enhances task-related features that are learned in an implicit manner. VSAM~\cite{quintino2019pose} uses the inference results of pose detection for supervision to train a pose-guided attention model. Although attention mechanisms are well-known methods for boosting performance, such mechanisms mainly discuss where to focus given spatial regions. Because we consider attention mechanisms as an extension that can be applied to any existing method with slight modifications, such mechanisms are given little attention in this work. One of main challenges in VAP is solving multi-task learning problems in the presence of label sparsity and class imbalances in training data~\cite{abdulnabi2015multi}. A single unified model is preferred for saving on the cost of inference and is expected to provide robust predictions for all targets (\textit{e.g., style, texture, and patterns in fashion}). The most commonly used method is to train multiple binary classifiers~\cite{liu2016deepfashion,corbiere2017leveraging}. However, because the numbers of effective annotations for each task differ significantly based on label sparsity, models can easily encounter overfitting on tasks with many annotations and underfitting on tasks with few annotations over the same number of iterations. Balancing such bias during training is important issue that must be solved for VAP. Adhikari \textit{et al.}~\cite{adhikari2019progressive} proposed a progressive learning approach that add branches for individual models for attributes progressively as training proceeds. Lu \textit{et al.}~\cite{lu2017fully} proposed a tree-like deep architecture that automatically widens the network for multi-task learning in a greedy manner. However, the implementation of such methods is tricky and requires considerable engineering work.

\noindent
\textbf{Label Embedding.}
Label embedding (LE) refers to an important family of multi-label classification algorithms that have been introduced in various studies~\cite{akata2015label,socher2013zero,palatucci2009zero}. Such approaches jointly learn mapping functions $\varphi$ and $\lambda$ that project embeddings of an image $x$ and label $y$ in a common intermediate space $Z$. Compared to direct attribute prediction, which requires training a single binary classifier for each attribute, LE has the following advantages. First, attributes are not required to be independent because one can simply move the embedding of $x$ closer to the correct label $y$ than to any incorrect labels $y'$ based on ranking loss ~\cite{hoffer2015deep,sohn2016improved}. Second, prediction classes can be readily expanded because classes can be predicted by measuring the shortest distance to the center point of a feature cluster assigned to an unseen class~\cite{kim2019edge,snell2017prototypical,khodadadeh2019unsupervised,ren2018meta}. We found a number of studies examining LE in terms of zero-shot learning. In this paper, we highlight the usage of LE to train domain experts for an MTSS model for semi-supervised learning.

\noindent
\textbf{Semi-supervised Learning and Distillation.}
Recent works~\cite{cevikalp2019semi,yalniz2019billion,xie2019self} have investigated semi-supervised learning (SSL) methods~\cite{van2019survey} that use large-scale unlabeled images to improve supervision. Yalniz \textit{et. al.}~\cite{yalniz2019billion} proposed a self-training method representing a special form of SSL and achieved state-of-the-art accuracy by leveraging a large amount of unannotated data. Distillation can also be considered as a form of SSL in that a teacher model makes predictions for unlabeled data and the results are used for supervision to train a student model~\cite{papernot2016semi,gong2018teaching,park2019relational}. Such strategies have yielded impressive results on many vision tasks. Inspired by these methods, we propose a training method that takes advantage of SSL for VAP. We found that a teacher-student paradigm of distillation is very effective for performing VAP because relevant visual attributes can be grouped into particular attribute types (\textit{e.g. $\{stripe, dot\} \in pattern$}), meaning a student model can effectively learn from multiple teachers that are specialists for each attribute type.

\section{Methodology}
\subsection{Overview}
The main concept of the MTSS approach is to integrate multiple teachers (MT) that are domain experts for each attribute type into a single student (SS) to construct a unified model that can predict multiple attributes in a single forward operation. Given an image $x$ and attribute type $\alpha_k \in \{ \alpha_1, \alpha_2, ... , \alpha_\kappa\}$, our goal is to predict the attribute classes $c_p \in \{ c^{\alpha_k}_1, c^{\alpha_k}_2,...,c^{\alpha_k}_P \}$, where $\kappa$ and $P$ are the number of target attribute types and final predictions to be outputted, respectively. Suppose $\alpha_k$ is a \textit{pattern} and \textit{color}. Then, the corresponding $c_p$ could be \textit{dot, stripe} or \textit{red}. $P$ may differ depending on which confidence score is used for the predicted result to reach a final decision regarding each attribute type $\alpha_k$. Our assumption is that attributes existing in the real world have a conceptual hierarchy, implying that relevant visual attributes are grouped into a particular attribute type $\alpha_k$. For example, $> 1000$ attributes in DeepFashion are grouped into 6 attribute types which means only 6 teachers are required for training student. The training procedure can be divided into two stages for the teacher and student. The design details and advantages of our two-stage training method are discussed in the following subsections.

\subsection{Teacher Models for Individual Attributes}
Given pairs of an image $x$ and ground-truth label $y$ in a training set $\zeta_{\alpha_k} = \{ (x_n, y_n), n=1,...,N\}$ for an attribute type $\alpha_k$ with $x_n \in X$ and $y_n \in Y$, our goal is to train a teacher model $\varphi^T_{\alpha_k}: X \rightarrow Y$. Our teacher model uses the pairwise ranking loss of metric learning to learn image representations. Specifically, a group of randomly initialized label embeddings obtained from an attribute dictionary $d^n_{\alpha_k} = \lambda_{\alpha_k}(y_n)$, where $d^n_{\alpha_k} \in R^D$ and $D$ denotes the dimension of the image embedding, is defined and $d^n_{\alpha_k}$ is learned to represent the center of the feature cluster of the $n$-th class for attribute type $\alpha_k$. Given a label embedding $d^n_{\alpha_k}$ that corresponds to the most representative feature point of the $n$-th class of $\alpha_k$, our goal is to locate an embedding of the positive image $e^+ = \varphi^T_{\alpha_k}(I^+)$ to be close to $d^n_{\alpha_k}$ and an embedding of the negative image $e^- = \varphi^T_{\alpha_k}(I^-)$ far away. The positive image is sampled from the same attribute class as $d^n_{\alpha_k}$ and the negative image is sampled from a randomly selected class that is different from the class of the positive image. Although various sampling strategies can be adopted to enhance overall performance~\cite{wu2017sampling}, such optimization was omitted in this work because it lies outside of our focus. The general form of an objective can be written as \eqref{eq:eq1}.
\begin{equation}
    \label{eq:eq1}
    \ell'(d^n_{\alpha_k},e^+,e^-) = max\{0, (1 - \lVert d^n_{\alpha_k} \rVert_2 \cdot \lVert e^+ \rVert_2 + \lVert d^n_{\alpha_k} \rVert_2 \cdot \lVert e^- \rVert_2)\}, \\
\end{equation}
where $\left \| \right \|_2$ represents L2 normalization. Lin \textit{et al.}~\cite{lin2017focal} introduced the concept of focal loss, which can help a model focus on difficult and misclassified examples by multiplying a factor $(1-p_t)^\gamma$ by the standard cross-entropy loss, where $p_t$ is the estimated probability for the target class. As a modification for metric learning, we applied focal loss to our method by defining the probability $p_t$ as the cosine similarity between $d$ and $e$, which is bounded in the range of $[0,1]$. The final objective is defined as \eqref{eq:eq2}.
\begin{equation}
    \label{eq:eq2}
    \begin{split}
    p_t = 0.5 \times max\{0, (1 + \lVert d^n_{\alpha_k} \rVert_2 \cdot \lVert e^+ \rVert_2 - \lVert d^n_{\alpha_k} \rVert_2 \cdot \lVert e^- \rVert_2)\}, \\
    \ell^{Teacher}(p_t) = - (1-p_t)^\gamma log(p_t), \quad \quad \quad \quad \quad \quad \\
    \end{split}
\end{equation}
where $\gamma$ is the hyperparameter to be found. Note that $\gamma$ was set to 1.0 in our experiments unless stated otherwise. We found that if $\gamma$ is adjusted carefully, adopting the modified focal loss term significantly boosts overall scores, which will be discussed in the experimental section. During training, an image can be duplicated in more than one attribute class if the multiple labels exist in the ground truth. We expect the number of experts (i.e., trained teacher models $\varphi^T \in \{\varphi^T_{\alpha_1}, \varphi^T_{\alpha_2}, ... ,\varphi^T_{\alpha_\kappa}\}$) to be the same as the number of attribute types $\kappa$.

\begin{figure}[t]
	\begin{center}
		\includegraphics[width=0.80\linewidth]{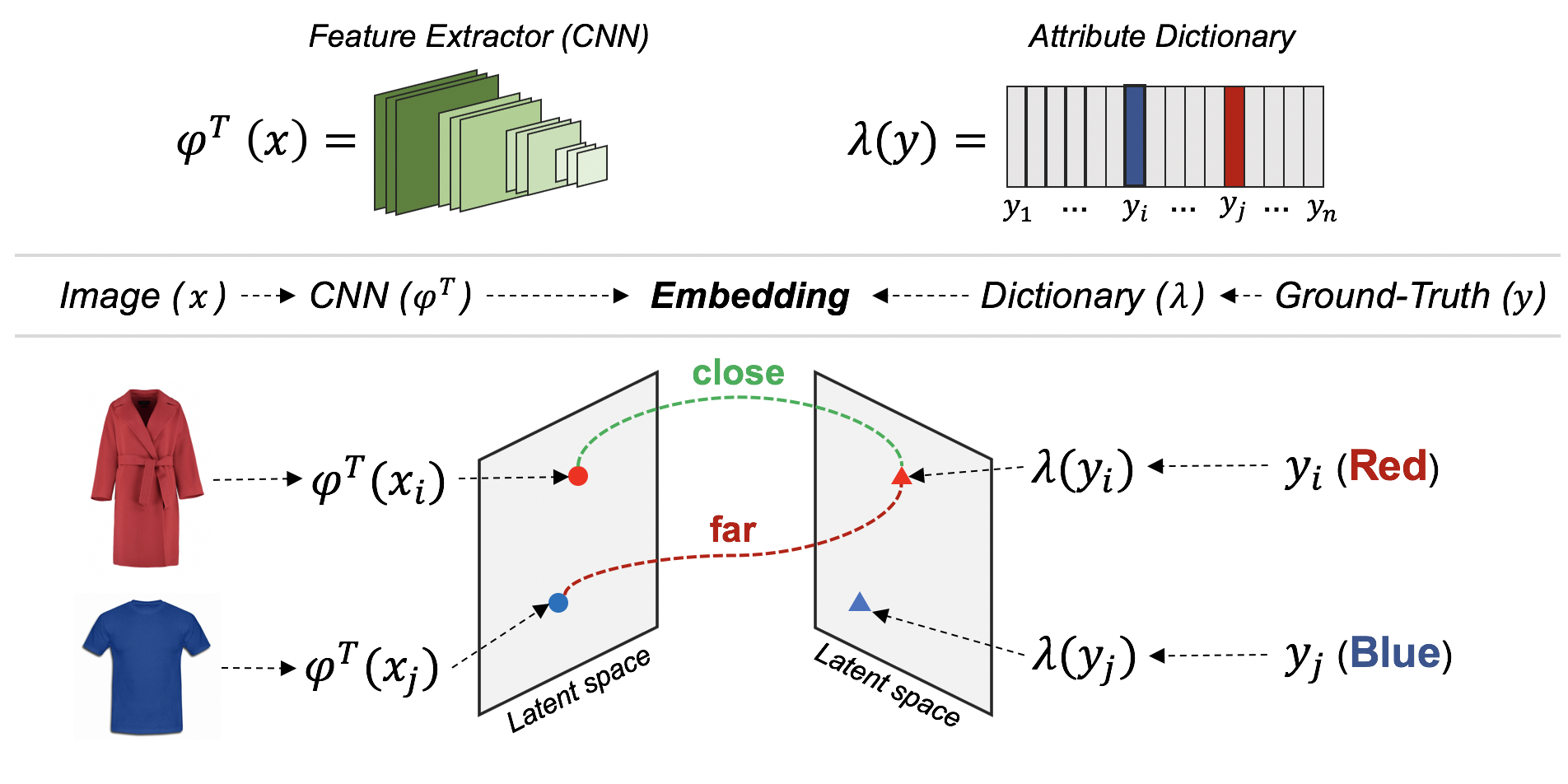}
	\end{center}
	\caption{Training of a teacher model. The feature extractor $\varphi^T$ and mapping $\lambda$ are learned to place the image embedding (left) and label embedding (right) close to each other in latent space.}
	\label{fig:training_teacher}
\end{figure}

\begin{figure}[t]
	\begin{center}
		\includegraphics[width=1.0\linewidth]{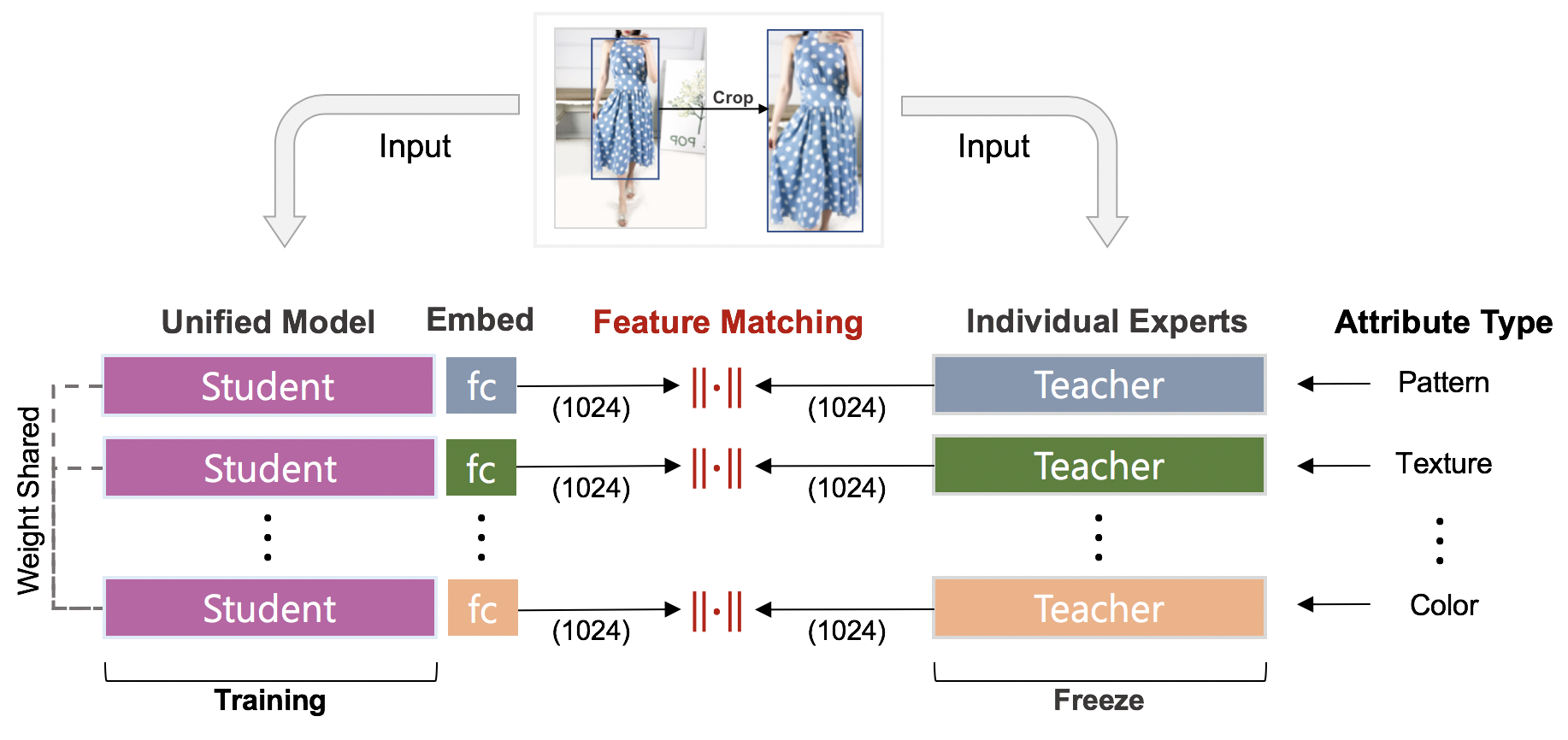}
	\end{center}
	\caption{Training of a student model. The teachers, which represent individual experts for each attribute type, are independent models, unlike the unified student model that shares weights before its fully connected layers.}
	\label{fig:training_student}
\end{figure}

\subsection{Unified Student Model for All Attributes}
The goal of the student stage is to integrate the trained teacher models $\varphi^T \in \{\varphi^T_{\alpha_1}, \varphi^T_{\alpha_2}, ... ,\varphi^T_{\alpha_k}\}$ into a single unified network $\varphi^S$ and boost overall performance by utilizing unlabeled images $\mathcal{U}$ in a semi-supervised manner. A distillation method that aims to transfer knowledge acquired in one model, namely a teacher, to another model, namely a student, was adopted for this purpose. The core idea of distillation is very simple. A student model is trained to mimic the feature distribution of $\varphi^T$. Learning distributions from trained models can be achieved simply by matching the image embedding of a teacher $e^T_{\alpha_k} = \varphi^T_{\alpha_k}(x)$ to the embedding of a student $e^S_{\alpha_k} = \varphi^S(x; \alpha_k)$ according to a target attribute type $\alpha_k$. It is assumed that if $\varphi^S$ is able to reproduce the same feature distribution as $\varphi^T$, then prediction can be performed by simply measuring the distance between $e^S_{\alpha_k}$ and a learned dictionary $d_{\alpha_k} \in \{d^1_{\alpha_k},d^2_{\alpha_k},...,d^N_{\alpha_k}\}$ that was already obtained in the teacher stage. A student $\varphi^S$ consists of $S$ branches of fully-connected layers following the last pooling layer of the backbone, where each branch is in charge of projecting pooled descriptors into an attribute-specific embedding space $Z_{\alpha_k} \in R^D$. The weights before the fully connected branches are shared. Given an attribute type $\alpha_k$, we maximize the cosine-similarity between $e^S_{\alpha_k}$ and $e^T_{\alpha_k}$. Therefore objective is formulated as

\begin{equation}
    \label{eq:eq3}
    \ell'' = \Sigma_{k=1}^\kappa \{ 1 - \lVert e^S_{\alpha_k} \rVert_2 \cdot \lVert e^T_{\alpha_k} \rVert_2 \}, \\
\end{equation}
where $\left \| \right \|_2$ is L2 normalization. Because $\varphi^S$ learns from $\varphi^T$, training can be unstable with large $lr$ values if $\varphi^T$ frequently produces outlier points based on a lack of generalization or excessive input noise. To alleviate such unwanted effects, we assign additional weight to an objective if the distance from the embedding of teacher $e^T_{\alpha_k}$ to the closest label embedding $\hat{d}_{\alpha_k}$ is small because a smaller distance indicates a more certain prediction. A large distance to $\hat{d}_{\alpha_k}$ could indicate the presence of outliers, so such signals are suppressed during gradient updating. The final objective is formulated as

\begin{equation}
    \label{eq:eq4}
    \ell^{Student} = \Sigma_{k=1}^\kappa \{ (\lVert \hat{d}_{\alpha_k} \rVert_2 \cdot \lVert e^T_{\alpha_k} \rVert_2)^{\beta} \times (1 - \lVert e^S_{\alpha_k} \rVert_2 \cdot \lVert e^T_{\alpha_k} \rVert_2) \}, \\
\end{equation}
where $\beta$ is the hyperparameter to be found. $\beta$ is set to 1.0 unless stated otherwise. Although we observed that $\beta$ has a very small effect on final accuracy, this setting enables to use a large $lr$ value at the very beginning of training phase by suppressing noise.

\subsection{Query Inference}
\begin{figure}[t]
	\begin{center}
		\includegraphics[width=1.0\linewidth]{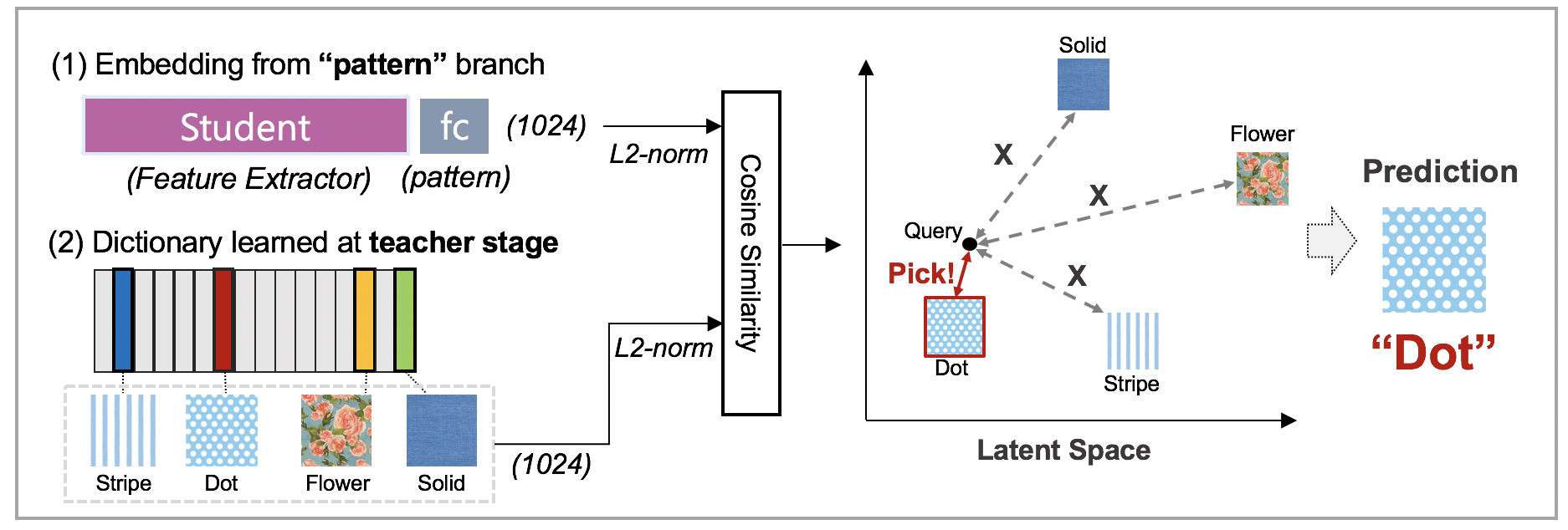}
	\end{center}
	\caption{An example of the query inference process.}
	\label{fig:query_inference}
\end{figure}

Given a query image $x$ and ground-truth label $y$ of an attribute type $\alpha_k$, the attribute-specific embedding of $\alpha_k$ can be calculated as $e^S_{\alpha_k} = \varphi^S_{\alpha_k}(x; \alpha_k)$. Let the dictionary learned at the teacher stage $d_{\alpha_k} = \lambda_{\alpha_k}(y)$, where $d_{\alpha_k} \in \{d^1_{\alpha_k}, ... ,d^N_{\alpha_k}\}$ and $N$ is the number of classes included in $\alpha_k$. Because $d_{\alpha_k}$ represents the center points of the feature cluster for each ground-truth class $y$ and $\varphi^S_{\alpha_k}$ can reproduce the same distribution with $\varphi^T_{\alpha_k}$, a prediction can be obtained by finding the $\hat{d}_{\alpha_k}$ that maximizes the cosine similarity with $e^S_{\alpha_k}$. An example of the query inference process for predicting a \textit{pattern} is presented in Figure~\ref{fig:query_inference}.

\section{Experiment}
\subsection{Experimental Setup}
\noindent
\textbf{Datasets.} All models were evaluated on several fashion-related benchmarks, namely three public datasets called iMaterialist-Fashion-2018 (iMatFashion)~\cite{guo2019imaterialist}, Deepfashion Category and Attribute Prediction Benchmark (DeepFashion)~\cite{liu2016deepfashion}, and DARN~\cite{huang2015cross}, and one private dataset called FiccY. We cropped the images using a CenterNet~\cite{zhou2019objects} based fashion object detector unless ground-truth bounding boxes were not provided for a dataset. We omit the detailed statistics of the public datasets because they have been well described in many related works~\cite{corbiere2017leveraging,wang2018attentive,guo2019imaterialist,shin2019semi}. FiccY is a private fashion dataset collected from our database and annotated by human experts. It contains 520K high-resolution images labeled for four types of attributes (\textit{Category, Color, Pattern, and Texture}) produced by both sellers and wearers. We intentionally include the result on FiccY in our experiments because it cab reflect a real service environment. However, the majority of experiments were performed on the DeepFashion benchmark for reproducibility.

\subsection{Performance on Benchmarks}

\noindent
\textbf{Multi-Label Classification.}
Previous studies ~\cite{chen2012describing,huang2015cross,liu2016deepfashion,corbiere2017leveraging,wang2018attentive,quintino2019pose,liu2018deep} on attribute prediction in the fashion domain have reported R@3 scores according to attribute types using DeepFashion, so we used the same metric for fair comparison. The results are listed in Table~\ref{tab:pure_performance}. The R@3 scores of both the teacher and student are listed for various experimental settings of $\gamma$ and $lr$. The results demonstrate that with the default hyperparameter of $\gamma=1$, our teacher model outperforms previous methods~\cite{corbiere2017leveraging,chen2012describing,huang2015cross} that use no auxiliary supervision. We determined that the optimized values of $\gamma$ and $lr$ differ depending on the task, meaning these hyperparameters must be selected carefully according to the target attribute type. When training a student, we observed higher gains when a greater $lr$ value was adopted. Adjusting the $\beta$ value enables the use of greater $lr$ values while avoiding instability at the beginning of training. The student trained by the tuned teachers yields very similar results compared to state-of-the-art domain-specific methods~\cite{liu2016deepfashion,wang2018attentive,quintino2019pose,liu2018deep}.

\begin{table}[ht]
\caption{Top-\textit{k} recall for attribute prediction on the DeepFashion~\cite{liu2016deepfashion} dataset. $\ast$-marked methods use additional labels for training, such as landmark, pose, or text descriptions. $\gamma=tune$ indicates that the best-performing $\gamma$ was selected according to the attribute types. Recall was measured when the model achieved the best F1@1 score. \textit{Style} is not compared to the other methods because style scores were not reproducible using publicly released code. Overall $1^{st}$/$2^{nd}$ best in \textcolor{blue}{\textbf{blue}}/\textcolor{green}{\textbf{green}}.}
\centering
\begin{adjustbox}{width=1\textwidth}
\begin{tabular}{ccccccccccccccc}
\hline
& \multicolumn{2}{c}{Category} & \multicolumn{2}{c}{Texture} & \multicolumn{2}{c}{Fabric} & \multicolumn{2}{c}{Shape} & \multicolumn{2}{c}{Part} & \multicolumn{2}{c}{Style} & \multicolumn{2}{c}{All} \\ \hline
Method & top-3 & top-5 & top-3 & top-5 & top-3 & top-5 & top-3 & top-5 & top-3 & top-5 & top-3 & top-5 & top-3 & top-5\\ \hline \hline
WTBI~\cite{chen2012describing} & 43.73 & 66.26 & 24.21 & 32.65 & 25.38 & 36.06 & 23.39 & 31.26 & 26.31 & 33.24 & - & - & 27.46 & 35.37 \\
DARN~\cite{huang2015cross} & 59.48 & 79.58 & 36.15 & 48.15 & 36.64 & 48.52 & 35.89 & 46.93 & 39.17 & 50.14 & - & - & 42.35 & 51.95 \\
$\ast$ FashionNet~\cite{liu2016deepfashion} & 82.58 & 90.17 & 37.46 & 49.52 & 39.30 & 49.84 & 39.47 & 48.59 & 44.13 & 54.02 & - & - & 45.52 & 54.61 \\
Corbiere et al.~\cite{corbiere2017leveraging} & 86.30 & 92.80 & 53.60 & 63.20 & 39.10 & 48.80 & 50.10 & 59.50 & 38.80 & 48.90 & - & - & 23.10 & 30.40 \\
$\ast$ Wang et al.~\cite{wang2018attentive} & \textcolor{green}{\textbf{90.99}} & \textcolor{green}{\textbf{95.78}} & 50.31 & \textcolor{green}{\textbf{65.48}} & 40.31 & 48.23 & 53.32 & 61.05 & 40.65 & \textcolor{green}{\textbf{56.32}} & - & - & 51.53 & 60.95 \\
$\ast$ VSAM + FL~\cite{quintino2019pose} & - & - & 56.28 & 65.45 & 41.73 & 52.01 & 55.69 & \textcolor{green}{\textbf{65.40}} & 43.20 & 53.95 & - & - & - & - \\
$\ast$ Liu et al.~\cite{liu2018deep} & \textcolor{blue}{\textbf{91.16}} & \textcolor{blue}{\textbf{96.12}} & 56.17 & \textcolor{blue}{\textbf{65.83}} & 43.20 & \textcolor{blue}{\textbf{53.52}} & \textcolor{blue}{\textbf{58.28}} & \textcolor{blue}{\textbf{67.80}} & 46.97 & \textcolor{blue}{\textbf{57.42}} & - & - & - & - \\
\hline
Ours(T)($\gamma=0$) & 88.45 & 91.21 & 55.91 & 62.02 & 43.90 & 50.54 & 55.92 & 61.46 & 45.40 & 51.55 & 30.13 & 35.44 & - & - \\
Ours(T)($\gamma=1$) & 89.32 & 92.92 & 57.89 & 64.71 & 44.51 & 51.57 & 56.22 & 62.34 & 46.09 & 52.73 & 30.88 & 36.19 & - & - \\
Ours(T)($\gamma=tune$) & 90.01 & 93.74 & \textcolor{green}{\textbf{58.14}} & 65.29 & \textcolor{green}{\textbf{44.93}} & 52.18 & 56.74 & 62.66 & 46.83 & 53.66 & 31.31 & 37.13 & - & - \\
Ours(S)($\gamma=1,\beta=1$) & 89.65 & 93.29 & \textcolor{blue}{\textbf{58.33}} & 65.17 & \textcolor{blue}{\textbf{45.05}} & 52.26 & 56.91 & 63.17 & \textcolor{green}{\textbf{46.98}} & 54.30 & \textcolor{green}{\textbf{32.30}} & \textcolor{green}{\textbf{38.41}} & \textcolor{green}{\textbf{57.85}} & \textcolor{green}{\textbf{63.91}} \\
Ours(S)($\gamma=tune,\beta=1$) & 90.17 & 93.98 & 57.58 & 64.66 & 44.87 & \textcolor{green}{\textbf{52.47}} & \textcolor{green}{\textbf{57.34}} & 63.69 & \textcolor{blue}{\textbf{47.36}} & 54.99 & \textcolor{blue}{\textbf{32.64}} & \textcolor{blue}{\textbf{39.24}} & \textcolor{blue}{\textbf{58.02}} & \textcolor{blue}{\textbf{64.35}} \\
\hline
\end{tabular}
\end{adjustbox}
\label{tab:pure_performance}
\end{table}

\noindent
\textbf{LE versus Binary Cross-Entropy}
In this section, we analyze the strength of LE compared to the most commonly used method, namely binary cross-entropy (BCE).
Table~\ref{tab:df_preds} compares our implementation to the existing method. The student model is trained by the teacher networks, which are trained using LE. The state-of-the-art method based on DeepFashion uses BCE for training.
One can see that LE consistently yields better results when \textit{k}=1, implying that prediction is more accurate with a small number of trials.
We directly compared the LE and BCE methods by replacing the objective of our teacher model with BCE.
The results in Table~\ref{tab:bce_vs_le} reveal consistent improvements in terms of recall at \textit{k}=1 for all evaluation datasets.

\begin{figure}[t]
	\begin{center}
		\includegraphics[width=1.0\linewidth]{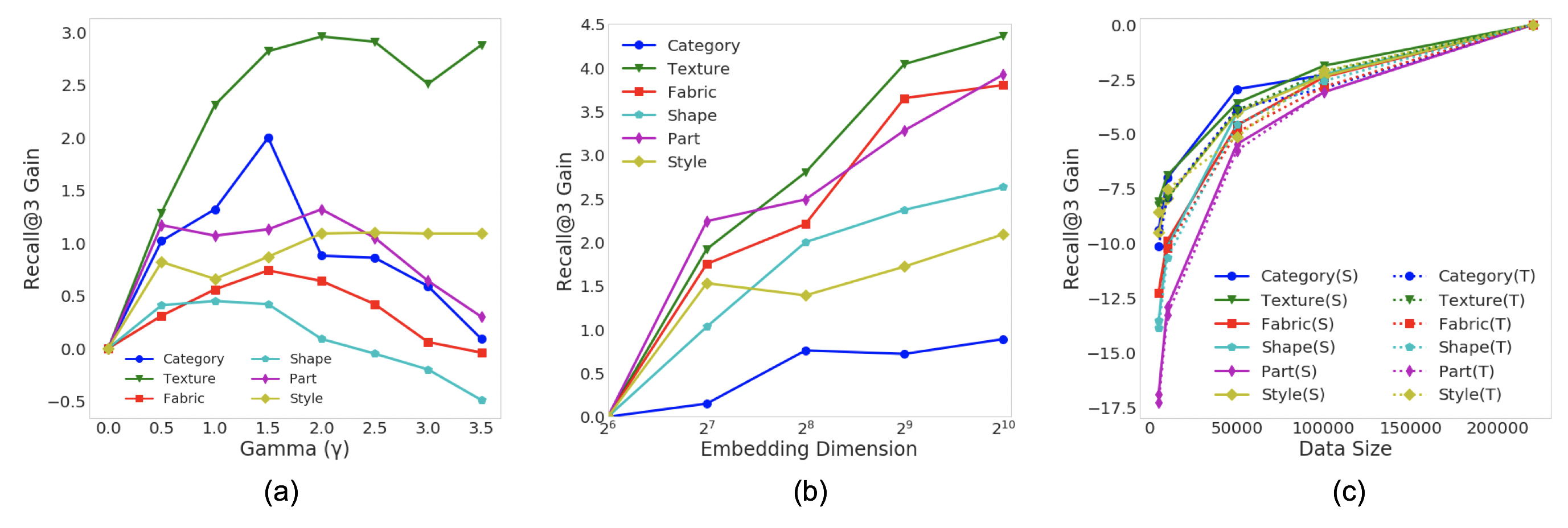}
	\end{center}
	\caption{Effects of training hyperparameters evaluated on DeepFashion: (a) gamma $\gamma$, (b) embedding dimension $D$, and (c) training data size. These plots represent analysis of the changes in R@3 (gain or degradation) compared to the baseline trained with (a) $\gamma=0$, (b) $D=256$, and (c) a data size equal to the size of the entire training set. Part (c) presents results for both the student (S) and teacher (T).}
	\label{fig:hyper_exp}
\end{figure}

\begin{table}[b]
\caption{The top-\textit{k} F1 scores and recall values evaluated on DeepFashion.}
\centering
\begin{adjustbox}{width=0.75\textwidth}
\begin{tabular}{ccccccc}
\hline
Method & F1@1 & R@1 & F1@3 & R@3 & F1@5 & R@5 \\ \hline \hline
Liu et al.~\cite{liu2018deep} & 31.35 & 37.17 & \textbf{22.50} & 57.30 & \textbf{16.82} & \textbf{65.66} \\
\hline
Ours(S)($\gamma=tune,\beta=1$) & \textbf{33.53} & \textbf{40.11} & 22.41 & \textbf{58.02} & 16.16 & 64.35 \\
\hline
\end{tabular}
\end{adjustbox}
\label{tab:df_preds}
\end{table}

\noindent
\textbf{Hyperparameter tuning.}
The optimal value of $\gamma$ for training a teacher network is analyzed in Figure~\ref{fig:hyper_exp} (a). While the use of focal loss consistently improves the results by up to 3\% in terms of R@3, we found that the optimal value differs depending on the attribute types. Therefore, it is highly recommended to tune $\gamma$ carefully for each attribute type. The feature dimension is also an important engineering factor. Figure~\ref{fig:hyper_exp} (b) presents the effects of the embedding dimension $D$. Better results can be observed for higher dimension, but processing high-dimensional features may require more resources and inference time. The result in Table~\ref{tab:pure_performance} demonstrate that tuning the hyperparameters of not only $\gamma$ and $D$, but also $lr$, significantly improves performance. The combination of optimal values is different for each attribute type.

\begin{table}[t]
\caption{Comparison between LE and BCE. The F1@1 scores were measured on FiccY.}
\centering
\begin{adjustbox}{width=0.6\textwidth}
\begin{tabular}{cccccc}
\hline
Dataset & Category & Pattern & Color & Texture & Avg. $\Delta$ \\ \hline \hline
BCE & 85.12 & 67.46 & 65.81 & 57.43 & -  \\
\hline
LE & \textbf{85.52} & \textbf{67.71} & \textbf{66.31} & \textbf{57.98} & \textcolor{green}{+0.62\%} \\
\hline
\end{tabular}
\end{adjustbox}
\label{tab:bce_vs_le}
\end{table}

\noindent
\textbf{Training depending on data size.}
Figure~\ref{fig:hyper_exp} (c) presents the performance degradation when the number of usable training images decreases. Overall R@3 scores are presented for both the teacher and student networks. The number of training images was intentionally limited to observe how performance degrades with a shortage of training data. As expected, a lack of training data significantly degrades the overall results. It is noteworthy that the student always outperforms the teachers slightly in terms of gain, even though the teachers are very weak. However, this also indicates that the final performance of the student is strongly bounded by that of teachers. We found no relationship between training data size and the amount of performance gain for a student compared to a teacher.

\begin{table}[b]
\caption{Performance comparison between a teacher (T) and student (S) model. F1@1 scores are measured on the iMatFashion, DeepFashion, and FiccY datasets.}
\parbox{1.0\linewidth}{
\centering
\begin{adjustbox}{width=0.75\textwidth}
\begin{tabular}{ccccccccc}
\hline
\multicolumn{9}{c}{iMatFashion~\cite{guo2019imaterialist}}                      \\
\hline
  & Category & Gender & Material & Pattern & Style & Neckline & Sleeve & Color \\
\hline \hline
T & \textbf{42.50} & 88.50 & 49.27 & 37.84 & 27.70 & 40.11 & 76.72 & 47.75   \\
S & 42.43 & \textbf{90.46} & \textbf{50.45} & \textbf{38.04} & \textbf{28.60} & \textbf{41.93} & \textbf{77.66} & \textbf{49.69}   \\
\hline
\end{tabular}
\end{adjustbox}
}
\\
\parbox{.5\linewidth}{
\centering
\begin{adjustbox}{width=0.53\textwidth}
\begin{tabular}{ccccccc}
\hline
\multicolumn{7}{c}{DeepFashion~\cite{liu2016deepfashion}}                      \\
\hline
  & Category & Texture & Fabric & Shape & Part & Style \\
\hline \hline
T & 74.45 & 25.61 & 24.97 & 30.45 & 21.12 & 12.63   \\
S & \textbf{74.85} & \textbf{26.08} & \textbf{25.14} & \textbf{30.81} & \textbf{21.40} & \textbf{13.29}  \\
\hline
\end{tabular}
\end{adjustbox}
}
\parbox{.5\linewidth}{
\centering
\begin{adjustbox}{width=0.40\textwidth}
\begin{tabular}{ccccc}
\hline
\multicolumn{5}{c}{FiccY}                      \\
\hline
  & Category & Pattern & Color & Texture \\
\hline \hline
T & 85.10 & 67.69 & 66.15 & 57.74   \\
S & \textbf{85.78} & \textbf{68.31} & \textbf{67.60} & \textbf{59.19}   \\
\hline
\end{tabular}
\end{adjustbox}
}
\label{tab:T_and_S_comparison}
\end{table}

\subsection{Effectiveness of Distillation}

\begin{figure}[t]
	\begin{center}
		\includegraphics[width=1.0\linewidth]{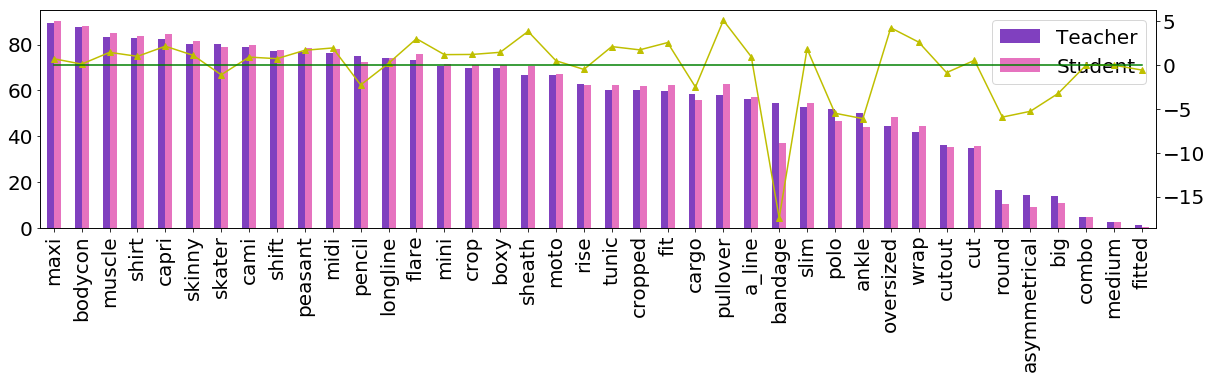}
	\end{center}
	\caption{R@3 scores for each class of attribute type \textit{shape} for the DeepFashion dataset. The yellow line represents the changes in performance before and after training the student. The green line is a baseline representing a change of zero percent.}
	\label{fig:shape_result}
\end{figure}

\noindent
\textbf{Comparison between a teacher and student.}
In Table~\ref{tab:T_and_S_comparison}, we compare classification performances between a teacher and student on the DeepFashion, iMatFashion, and FiccY datasets. Results were evaluated for all three datasets to determine if the proposed method exhibits consistent experimental trends in different domains. The results illustrate that the student achieves a better score than the teachers in all cases. Our interpretation is that the use of unlabeled images can induce improvements because the negative effects caused by missed annotations are successfully suppressed during training by matching features from teacher and student for supervision. As a proof, only slight improvements in terms of F1 score are observed for \textit{categories} which has relatively dense annotations than the others. Our MTSS approach is a semi-supervised form of multitask learning, so we analyze the benefits of learning representations simultaneously from multiple tasks in Table~\ref{tab:single_multi_branch}. Given four different attribute types for FiccY, we compare students trained with the supervision of either single or multiple domain experts. We found consistent improvements in both cases. However, higher improvement is achieved in general with supervision from all four teachers than a single teacher. We can conclude that the proposed model benefits from knowledge transfer between different tasks, which were represented by different attribute types in this experiment. Figure~\ref{fig:shape_result} analyzes the differences in R@3 scores between teachers and students for each class with $\alpha_\kappa = shape$. Although the result by class is generally improved based on distillation, performance degradation occurs in some classes. Such results are found when the recall of a teacher is poor meaning that poorly generalized teacher networks can negatively affect optimization for minor classes, while having a small effect on overall performance. Another possibility is that certain rare attributes lead to underfitting for certain classes, even with a large number of training epochs. Regardless, the overall results with distillation for all classes exhibit consistent improvement.

\begin{table}[t]
\caption{R@3 gain of a student model compared to teachers with either single (S-Single) or multiple (S-Multi) teachers evaluated on FiccY. $\Delta_{single}$ and $\Delta_{multi}$ denote the results for S-Single and S-Multi, respectively, as percentages.}
\centering
\begin{adjustbox}{width=0.78\textwidth}
\begin{tabular}{cccccc || c}
\hline
Attribute      & T     & S-Single & $\Delta_{single}$  & S-Multi & $\Delta_{multi}$ & $\Delta_{multi}$ - $\Delta_{single}$\\
\hline \hline
Category & 96.70 & 97.30    & \textcolor{green}{+0.62\%}  & \textbf{97.30}    & \textcolor{green}{+0.62\%}   &  \textcolor{green}{+0.00\%}  \\
\hline
Pattern  & 90.11 & 90.73    & \textcolor{green}{+0.69\%}  & \textbf{91.14}    & \textcolor{green}{+1.14\%}   &  \textcolor{green}{+0.45\%}  \\
\hline
Color    & 88.97 & 89.54    & \textcolor{green}{+0.64\%}  & \textbf{89.98}    & \textcolor{green}{+1.14\%}   &  \textcolor{green}{+0.50\%}  \\
\hline
Texture  & 92.06 & 93.57    & \textcolor{green}{+1.64\%}  & \textbf{94.22}    & \textcolor{green}{+2.35\%}   &  \textcolor{green}{+0.71\%}  \\
\hline
\end{tabular}
\end{adjustbox}
\label{tab:single_multi_branch}
\end{table}

\begin{figure}[ht]
	\begin{center}
		\includegraphics[width=0.90\linewidth]{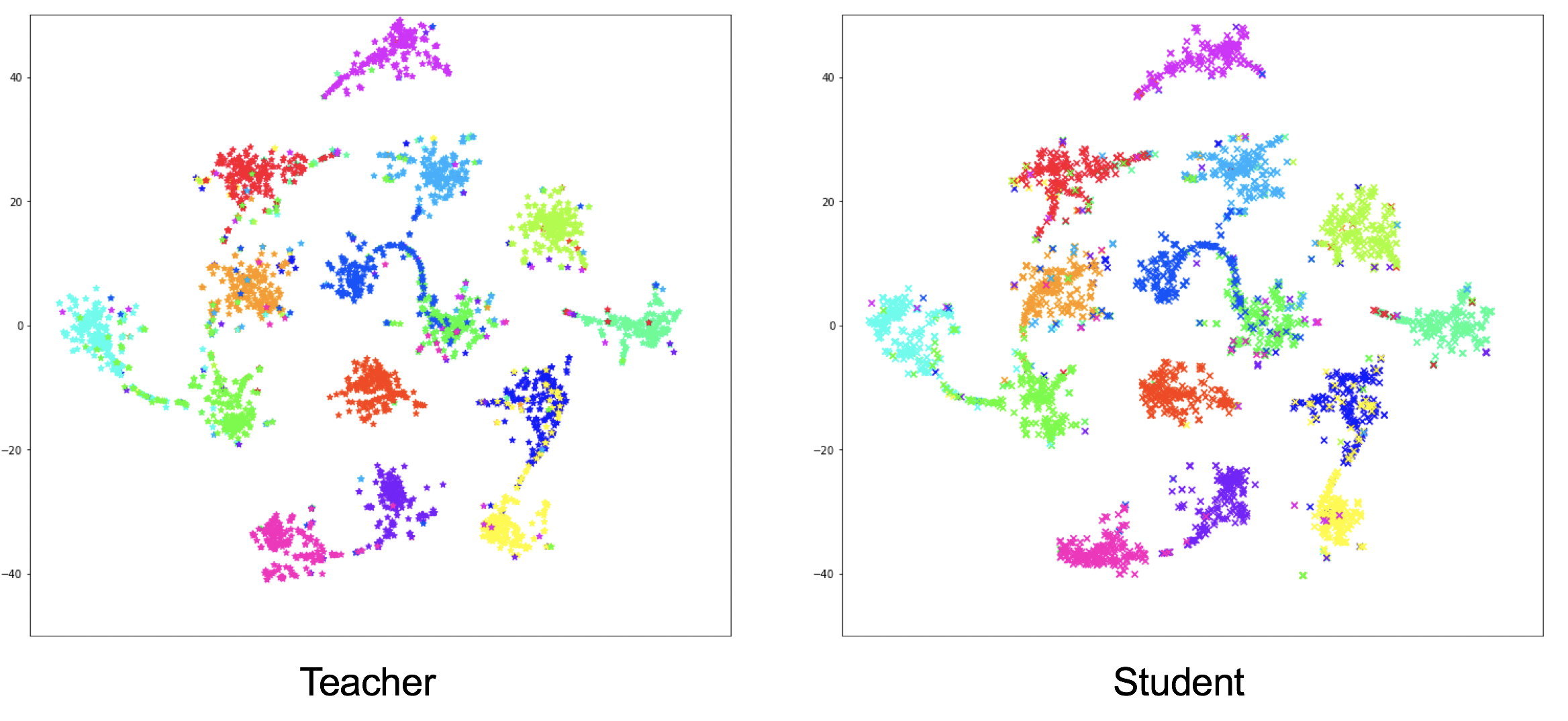}
	\end{center}
	\caption{T-distributed stochastic neighbor embedding (t-SNE)~\cite{maaten2008visualizing} visualization of pattern attribute types for FiccY. The left and right parts present the distributions of image embeddings extracted from the teacher and student models, respectively. }
	\label{fig:tsne}
\end{figure}

\noindent
\textbf{t-SNE visualization of distributions.}
The distribution of teacher and student embeddings is visualized in Figure~\ref{fig:tsne}. The result illustrates that the student reproduces almost the same distribution as the teacher. We omitted visualizations for attribute types other than \textit{patterns} because similar results were observed for all attribute types. The feature clusters for each prediction class are clearly formed for both the teacher and student.

\begin{table}[!h]
\caption{Cross-domain adaptability. F1@1 scores of five commonly appearing classes were measured on DARN~\cite{huang2015cross}. The symbol $^{\dag}$ indicates a model trained with both original training set images and external large-scale unlabeled images.}
\centering
\begin{adjustbox}{width=0.95\textwidth}
\begin{tabular}{ccccccccc}
\hline
Model  & Train set & Test set & Flower & Stripe & Dot & Check & Leopard & All \\
\hline
\hline
T  &  DeepFashion  &  DARN  &  70.46  &  66.11  &  67.47  &  51.00  &  40.90  &  64.31  \\
S  &  DeepFashion  &  DARN  &  71.16  &  67.81  &  68.97  &  52.02  &  43.97  &  65.15  \\
$S^\textsuperscript{\dag}$   &  DeepFashion+DARN  &  DARN  &  \textbf{71.84}  &  \textbf{68.70}  &  \textbf{70.03}  &  \textbf{52.62}  &  \textbf{44.09}  &  \textbf{65.49}  \\
\hline
T  &  iMatFashion  &  DARN  &  74.73  &  75.33  &  74.25  &  68.53  &  51.84  &  72.34  \\
S  &  iMatFashion  &  DARN  &  75.37  &  76.74  &  74.90  &  69.77  &  55.72  &  73.10  \\
$S^\textsuperscript{\dag}$  &  iMatFashion+DARN  &  DARN  &  \textbf{75.81}  &  \textbf{76.78}  &  \textbf{75.20}  &  \textbf{70.34}  &  \textbf{57.61}  &  \textbf{73.22}  \\
\hline
\end{tabular}
\end{adjustbox}
\label{tab:cross_domain}
\end{table}

\begin{table}[b]
\centering
\caption{Results for robustness to corruption and perturbation in inputs. Recall was measured on both clean and corrupted versions of the DeepFashion dataset. The symbol $\dagger$ indicates a model trained with both clean images and 1M, 10M, or 15M corrupted images.}
\begin{tabular}{cccccc}
\hline
Model        & Mixing & Clean R@3 & $\Delta$ & Corrupted R@3 & $\Delta$ \\
\hline \hline
Baseline(S)          & No           & 57.47          & -                          & 42.96             & -    \\
$S^{\dagger}$-1M     & Yes          & 57.75          & \textcolor{green}{+0.49\%} & \textbf{43.96}    & \textcolor{green}{+2.33\%} \\
$S^{\dagger}$-10M    & Yes          & \textbf{57.80} & \textcolor{green}{+0.57\%} & 43.88             & \textcolor{green}{+2.14\%} \\
$S^{\dagger}$-15M    & Yes          & 57.74          & \textcolor{green}{+0.47\%} & 43.84             & \textcolor{green}{+2.05\%} \\
\hline
\end{tabular}
\label{tab:robustness}
\end{table}

\subsection{Robustness and Generalization}

\noindent
\textbf{Cross-domain adaptability.}
In this subsection, we examine the capability of SSL to transfer knowledge for adaptation to a new target domain~\cite{orbes2019knowledge}. Validating the cross-domain adaptability of VAP is difficult because the classes provided in each dataset differ. For fair comparison, we carefully selected five commonly appearing classes (\textit{Flower, Stripe, Dot, Check, and Leopard}) for three datasets (\textit{DeepFashion, iMatFashion, and DARN}) and created subsets containing only images from one of the selected classes. The results in Table~\ref{tab:cross_domain} reveal that training a student is effective for cross-domain classification. We observe clear improvements for all five classes relative to the teacher models. We also examined the strength of SSL useful for deploying a model in a realistic environment. Because the training process for a student requires no annotations, the training images for two distinct domains (DARN and DeepFashion/iMatFashion) were mixed for training. This combination provided a significant performance boost for the testing set. This could be an useful feature in real production environments because a model can be easily fine-tuned for a service domain after being trained in a constrained environment.

\noindent
\textbf{Robustness to corruption and perturbation.}
Hendrycks \textit{et al.}~\cite{hendrycks2019benchmarking} introduced a benchmark for image classifier robustness to corruption and perturbation. Images were transformed using 15 types of algorithmically generated corruptions with five levels of severity, resulting in a total of 15 $\times$ 5 corrupted images. By following this same procedure, we created a corrupted version of the DeepFashion dataset and measured the drop in R@3 compared to the clean version. The result are listed in Table~\ref{tab:robustness}. Testing on the corrupted images degrades the overall R@3 score by an average of approximately 25\%. In Table~\ref{tab:cross_domain}, the benefits of mixing two distinct domains in terms of domain adaptation were explored. Following the same strategy, we mixed clean images with 1M, 10M, and 15M randomly sampled corrupted images and trained a student model to improve robustness to corruption. It should be noted that the mixed images were only used for training the student, meaning the teachers never saw the corrupted images. The results reveal that a student trained with mixed images achieves an increase of up to 2.33\% in terms of R@3 score compared to a baseline student trained using only clean images. Additionally, performing fine-tuning using corrupted images does not degrade the model’s performance on clean images. In fact, such fine-tuning slightly improves performance on clean images. The greatest performance gain can be observed when including only 1M corrupted images. The optimal ratio between clean and corrupted images should be identified because the noisy signals generated by corrupted images can make optimization less effective if they dominate the clean images.

\section{Conclusion}
In this paper, we proposed an MTSS approach to solving the VAP problem. Our method trains a unified model according to multiple domain experts, which enables it to predict multiple attributes that appear simultaneously in objects using a single forward operation. The core idea of MTSS is to transfer knowledge by forcing a student to reproduce the feature distributions learned by teachers. We demonstrated that such a strategy is highly effective for VAP, which suffers from a lack of effective annotations. Furthermore, our method can achieve competitive results on benchmarks using attribute labels alone, and improve robustness and domain adaptability without sacrificing accuracy fine-tuning with unlabeled images.

\newpage

\bibliographystyle{splncs}
\bibliography{paper}

\end{document}


\pagestyle{headings}
\mainmatter
\def\ECCVSubNumber{1381}  

\appendix
\section{Dataset overview}
\vspace{-7mm}
\begin{figure}[h]
	\begin{center}
		\includegraphics[width=0.95\linewidth]{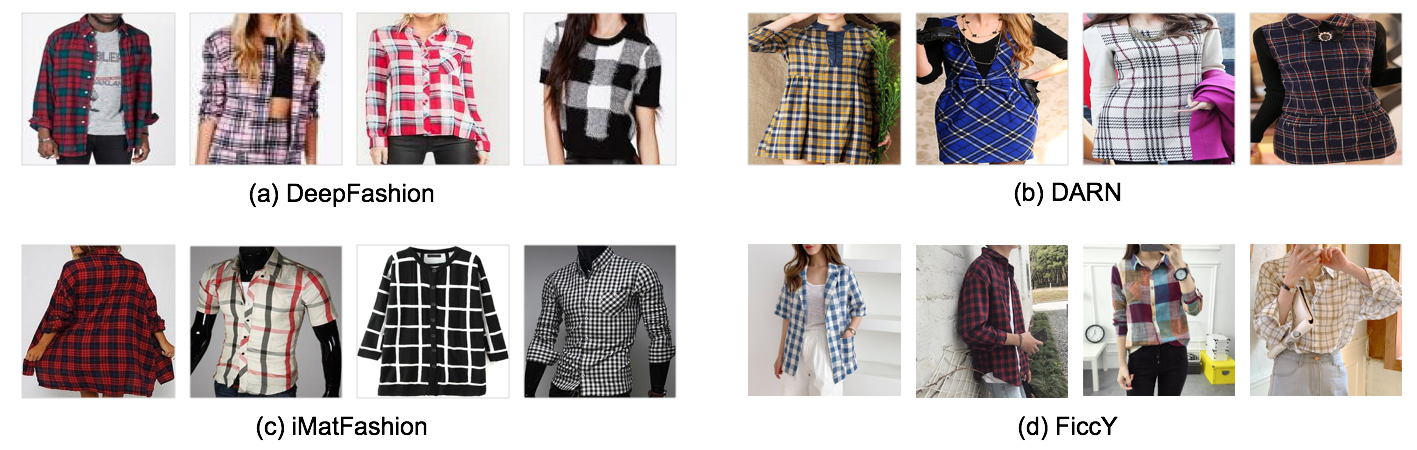}
	\end{center}
	\vspace{-4mm}
	\caption{Samples of ground-truth "check" in each benchmark datasets (DeepFashion, DARN, iMatFashion and FiccY). The fact the quality of the image (\textit{e.g. view point, background noise etc.}) significantly differs according to the dataset highlights the importance of domain adaptability. }
\end{figure}

\section{Effect on training the student varying the amount of unlabeled data }

\begin{table}[ht]
\caption{The comparison of mean AP (mAP) between the teacher (T) and the student (S) varying the amount of unlabeled images used for training.}
\centering
\begin{tabular}{c||c||cccccc}
\hline
mAP           & T     & \multicolumn{6}{c}{S (Multi-task learning with 4 teachers)} \\
\hline
Train Size    & -     & 1K         & 5K        & 10K       & 50K       & 100K      & 500K      \\
\hline
Category      & \textbf{85.50}  & 45.38 & 64.08 & 73.76 & 84.43 & \textcolor{green}{\textbf{86.04}} & \textcolor{green}{\textbf{87.18}}     \\
Pattern       & \textbf{72.54}  & 13.11 & 41.46 & 57.80 & 70.72 & \textcolor{green}{\textbf{72.9}} & \textcolor{green}{\textbf{73.72}}     \\
Color         & \textbf{58.58}  & 14.54 & 50.60 & 54.12 & 57.76 & 58.48 & \textcolor{green}{\textbf{59.33}}     \\
Texture       & \textbf{62.30}  & 40.90 & 50.61 & 56.57 & \textcolor{green}{\textbf{62.97}} & \textcolor{green}{\textbf{63.75}} & \textcolor{green}{\textbf{64.20}}      \\
\hline
\end{tabular}
\end{table}

\section{Evaluation Metrics}
The proposed model was mainly evaluated using the two most frequently used metrics: recall and F1 score. Recall is also referred to as sensitivity and it measures the probability of a positive detection. Precision, which is frequently used in combination with recall, is the percent of all relevant results among returned predictions. The F1 score is the harmonic mean of prediction and recall. The F1 score can fluctuate depending on which confidence score is used for a class to be considered as a final prediction. To minimize misleading effect by thresholding strategy, predictions were sorted by score and the top-\textit{k} classes were selected as final predictions, meaning the number of predictions for an image was always $k$.

\newpage

\section{Example of corruptions in DeepFashion}
\vspace{-7mm}
\begin{figure}[h]
	\begin{center}
		\includegraphics[width=0.65\linewidth]{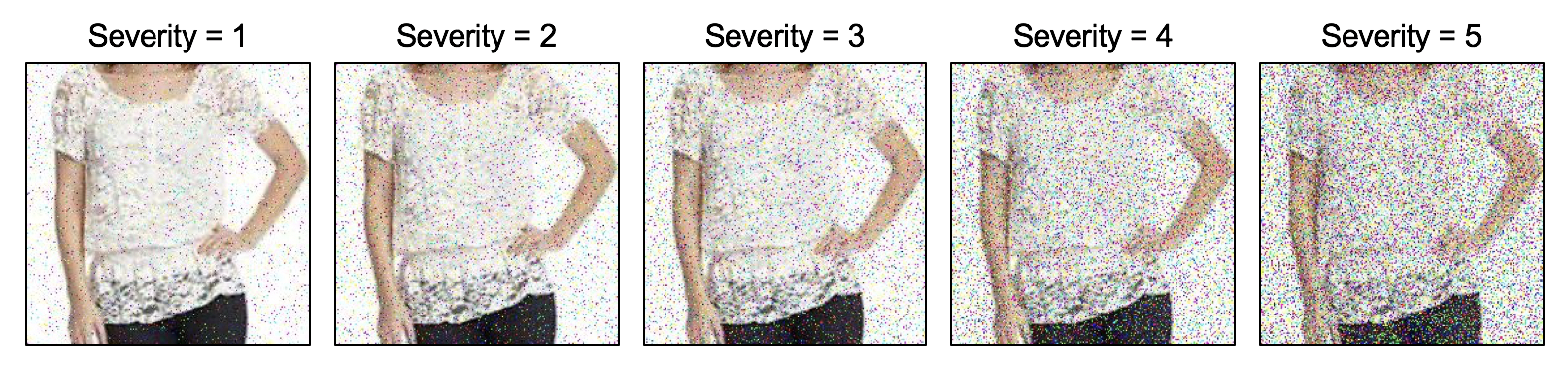}
	\end{center}
	\vspace{-14mm}
\end{figure}

\begin{figure}[h]
	\begin{center}
		\includegraphics[width=0.65\linewidth]{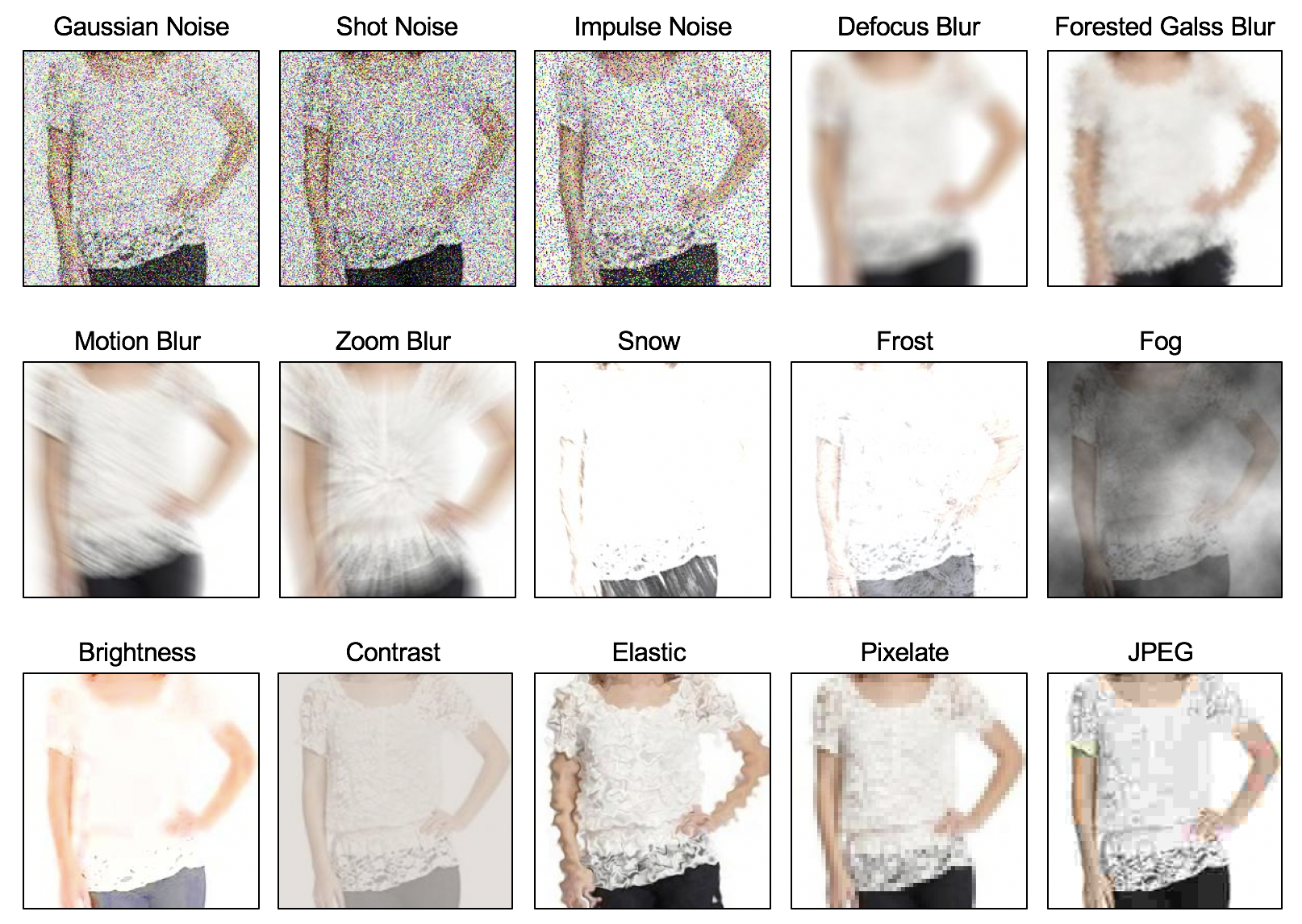}
	\end{center}
	\vspace{-4mm}
\end{figure}

\section{Implementation details}

\noindent
\textbf{General.}
A ResNet50 was used as a backbone for both the teacher and student networks. The number of output dimensions for image embedding was set to 1024. The images for each class were randomly sampled to collect 300K images for each epoch of training. Images were re-sampled for every epoch. A stochastic gradient descent optimizer with default parameters provided by PyTorch was adopted. The classes were sorted by the number of images and only images included in top-40 classes were used for training, while evaluation was performed using all classes. Our implementation is based on PyTorch and the proposed model was converted into the Open Neural Network Exchange for deployment for real production.

\noindent
\textbf{Teacher stage.}
The initial learning rate $lr$ was set to 0.2 with a batch size of 128. $lr$ was set to decay by a factor of 0.5 at every 10 epochs over 40 total epochs. $\gamma$ was set to 1.0 by default. The images with no ground-truth label for the target attribute type were not used for training the teacher to avoid the unnecessary computational overhead.

\noindent
\textbf{Student stage.} $lr$ was gradually increased from 0 to 0.4 at the beginning until 1M images were processed. $lr$ was set to decay by a factor of 0.5 at every 10 epochs over 100 total epochs. $\beta$ was set to 1.0 by default.